\documentclass[conference]{IEEEtran}
\IEEEoverridecommandlockouts
\usepackage{cite}
\usepackage{amsmath,amssymb,amsfonts}
\usepackage{algorithmic}
\usepackage{graphicx}
\usepackage{textcomp}
\usepackage{xcolor}
\usepackage{url}
\usepackage{subfigure}
\usepackage{multirow}
\usepackage{adjustbox}
\usepackage{hyperref}
\hypersetup{hidelinks,
	colorlinks=true,
	allcolors=black,
	pdfstartview=Fit,
	breaklinks=true}

\newtheorem{definition}{Definition}
\def\BibTeX{{\rm B\kern-.05em{\sc i\kern-.025em b}\kern-.08em
    T\kern-.1667em\lower.7ex\hbox{E}\kern-.125emX}}
\begin{document}

\newcommand{\note}[1]{\textcolor{blue}{[***#1***]}}


\title{\huge{Expert Knowledge-Guided Length-Variant Hierarchical \\Label Generation for Proposal Classification}
\thanks{This work was supported by Natural Science Foundation of China under Grant No. 61836013,
Beijing Natural Science Foundation under Grant No. 4212030,
Beijing Nova Program of Science and Technology under Grant No. Z191100001119090,
Science and Technology Service Network Initiative, Chinese Academy of Sciences (No. KFJ-STS-QYZD-2021-11-001) and 
Youth Innovation Promotion Association CAS.
}}

\author{\IEEEauthorblockN{ Meng Xiao$^{1,2,\S}$, Ziyue Qiao$^{1,2,\S}$\thanks{$\S$These authors have contributed equally to this work.}, Yanjie Fu$^3$,Yi Du$^{1,*}$\thanks{$^*$Corresponding author.},Pengyang Wang$^4$, Yuanchun Zhou$^1$}
\IEEEauthorblockA{\textit{$^1$Computer Network Information Center, Chinese Academy of Sciences, Beijing} \\
\textit{$^2$University of Chinese Academy of Sciences, Beijing} \\
\textit{$^3$Department of Computer Science, University of Central Florida, Orlando}\\
\textit{$^4$University of Macau, Macau}\\
\{shaow, qiaoziyue\}@cnic.cn, yanjie.fu@ucf.edu,
pywang@um.edu.mo, \{duyi, zyc\}@cnic.cn}}
\maketitle

\begin{abstract}

To advance the development of science and technology, research proposals are submitted to open-court competitive programs developed by government agencies (e.g., NSF). Proposal classification is one of the most important tasks to achieve effective and fair review assignment. 
Proposal classification aims to classify a proposal into a length-variant sequence of labels.
In this paper, we formulate the proposal classification problem into a hierarchical multi-label classification task. 
Although there are certain prior studies, proposal classification exhibit unique features:  
1) the classification result of a proposal is in  a hierarchical discipline structure with different levels of granularity;
2) proposals contain multiple types of documents; 
3) domain experts can empirically provide partial labels that can be leveraged to improve task performances. 
In this paper, we focus on developing a new deep proposal classification framework to jointly model  the three features. 
We design a deep  transformer-based encoder-decoder framework. 
In this framework, we use a two-level (word-level and document-level) Transformer structure  as an encoder to learn the embedding feature vectors of proposals. 
The decoder generates labels from the starting coarse-grained level to a certain fine-grained level to form the hierarchical discipline tree. In particular, to sequentially generate labels, we leverage previously-generated labels  to predict the label of next level; to integrate partial labels from experts, we use the embedding of these empirical partial labels to initialize the state of neural networks. Our model can automatically identify the best length of label sequence to stop next label prediction.  
Finally, we present extensive results to demonstrate that our method can jointly model partial labels, textual information, and semantic dependencies in label sequences and, thus, achieve advanced performances.  
\end{abstract}


\setlength{\baselineskip}{11.3pt}

\section{Introduction}
%

Today, research funding is awarded based on the intellectual, education, socio-societal merits of proposals. Therefore, proposals are submitted to open-court competitive programs managed by government agencies (e.g., NSF). Later, proposals are assigned to appropriate reviewers to solicit review comments and ratings. One of the pains of running such peer-review system is to assign a proposal to an appropriate panel and a set of appropriate reviewers, so as to advance the effectiveness and fairness of the review process. To achieve this goal, it is critical to accurately classify a proposal into a correct label of its discipline and topic. 
In this paper, we study the problem of proposal classification that aims to classify a proposal into a length-variant sequence of labels.

With the exponentially increasing number of universities, faculty members, and graduate student hiring, the submissions of research proposals has been exploding.
For example, the Natural Science Foundation of China (NSFC) has to manage (i.e., categorization, organizing panels, finding reviewers, selecting awardees, post award managements) more than one million research proposals every year.  There is an urgent need to bring in AI to assist with proposal categorization, review assignment, panel discussions. However, it is challenging to teach a machine to understand an unstructured proposal and categorize the proposal due to the unstructureness of the proposal.  However, is the information unstructureness of a proposal always unsolvable barrier in proposal classification? How can we discover and model structured knowledge from unstructured proposal data?


After analyzing large-scale proposal data from NSFC, we identify three important unique properties of  proposal data. These unique properties provide great potential to introduce structured knowledge into proposal classification models. 

\begin{figure}
\includegraphics[width=0.45\textwidth]{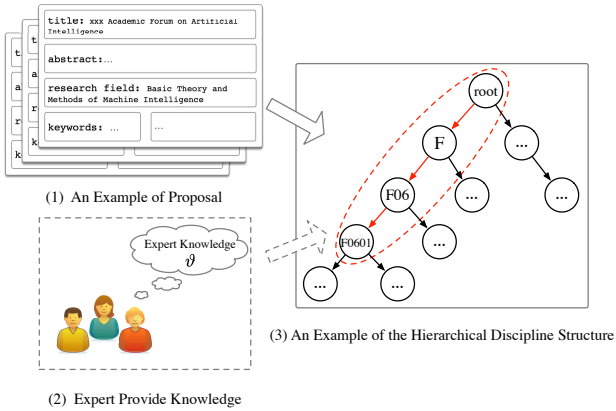} 
\caption{A Toy Model of Proposal Classification.}
\label{Fig.label_1}
\end{figure}

First, proposals exhibit a \textbf{hierarchical discipline label structure}. As known to all, a discipline system include many discipline labels  and subdiscipline labels. For example, in NSFC, the disciplines are organized as a taxonomy system, which we call ApplyID\footnote{the details of ApplyID code is list in here:  \href{http://www.nsfc.gov.cn/publish/portal0/tab550/}{[link]}.}. 
This system contains thousands of disciplines and subdisciplines with different levels of granularity and exhibits a hierarchical structure. The ApplyID system is viewed as a Directed Acyclic Graph(DAG) or a tree. Every ApplyID code is prefixed by a capital letter from A to H, representing a discipline or a subdiscipline, followed by zero- to six-digit. Every two-digit number in the ApplyID code represents a subdiscipline division in a particular granularity. For example, in Figure \ref{Fig.label_1}(3), F refers to the main discipline \textit{Information Sciences}. F06 represents \textit{Artificial Intelligence}, a subdiscipline of \textit{Information Sciences}, and F0601 represents \textit{Fundamentals of Artificial Intelligence}, a subdiscipline of F06. Thus, there is a hierarchical structure underneath the proposal classification results.  More importantly, the classification labels of a proposal can be  represented as a sequential discipline path through a tree structure. The paths can have arbitrary length as different proposal may cover different granularity level of disciplines, which we call a length-variant hierarchical discipline label path.

Second, in real world practice, domain experts usually provide \textbf{external partial labels} to a proposal. These external labels are usually more accurate and based on empirical experience. 
If a domain expert label a proposal with a discipline category in upper levels, can we treat these partial labels given by external experts as a kind of external knowledge to guide the proposal classification? If properly leveraged, we can advance the label classification accuracy. For example, Figure \ref{Fig.label_1} shows it is possible to utilize partially available labels as external expert knowledge to find the optimized fine-grain label path on the hierarchical discipline system for a proposal. In addition,  as the hierarchical level depth increases, the number of disciplines or subdisciplines will increase rapidly, making the prediction of next discipline labels very difficult. 
Finally, the textual data of a proposal can be decomposed into and analogized as \textbf{a mixture of multi-type documents}. For example, Figure \ref{Fig.label_1}(1) shows a proposal can be treated as a mixture of multiple types of documents (e.g., title, abstract, main body, keywords...). In addition, the semantic categories of title, abstract, main body are critical in assigning labels to the proposal. How can we leverage such internal multi-type document structure to enhance proposal classification? 


To jointly model the three unique features of proposal data, we formulate this proposal classification problem  into a  hierarchical multi-label classification task.  Different from classic hierarchical multi-label classification, our machine learning task is subject to the following constraints: 1) labels are related in a hierarchical fashion. 2)  the length of label sequence is unknown. So, our model need to automatically infer the optimized length of the label sequence output by the proposal classifier. 3) our model should have the ability to integrate external partial labels. 4) our model should have the ability to model the internal structure (multi-type documents: title, abstract, body) of a proposal.  

To this end, we propose a \textbf{H}ierarchical \textbf{M}ulti-label proposal classification method via \textbf{T}ransformer based encoder-decoder (HMT), which can assign the proposal to a correct path of disciplines on a hierarchy structure. Specifically, we first propose a two-level hierarchical encoder to fully capture both the documents' semantic information and the type information, then aggregate it into latent representations to representing the proposal. Then, we use a Transformer-based decoder to perform hierarchical multi-label classification. The prediction can start from the root label or any given previous label, which is incorporated with the representation of the proposal via the decoder to generate the discipline label in sub-level. On top of that, The decoder takes the previously generated label into account to generate the next label to hold the dependency between each level. The model generates the labels iteratively until achieving a proper level on the discipline hierarchy. The generated discipline label represents the most appropriate fine-grained label of the proposal. In conclusion, our contributions are as follows: 
\begin{itemize}
    \item We explore and formulate the problem of proposal classification on hierarchical discipline structure. We model this problem as a hierarchical label generation task, which can utilize the expert knowledge and predict the disciplines of the proposal with a appropriate level of granularity.
    \item We propose a Transformer based encoder-decoder framework. The encoder including two-level Transformers can extract abundant textual information from a mixture of multi-type proposal documents. The decoder can learn the dependency between labels on the hierarchical discipline structure, and output the length-variant label sequence via a top-down generation process.
    \item We conduct experiments on a real-world dataset to prove the effectiveness and explanatory power of our model. All the results from our experiment demonstrate that the model we propose can solve the real-world problems. 
\end{itemize}
\section{PRELIMINARIES}
In this section,  we start by giving the definition to the proposal and hierarchical discipline structure mentioned before. Then, we give a formal definition of the Hierarchical Multi-label Proposal Classification(HMPC) problem.
\begin{definition}[\textbf{Proposal}]
Like the toy example in Figure \ref{Fig.label_1}(1), a proposal may have multiple textual characteristics, including title, abstract, and keywords. We collect the text information of these characteristics as documents and orderly compose these documents of each proposal together to a  document set.
Formally, let $\mathbb{A}=\{D_1, D_2, D_3,...,D_{T}\}$ be a proposal which is composed of a set of documents with $T$ different types. Every document $D_i$ in the proposal, denoted as $D_i=[w^{(i)}_1,w^{(i)}_2,...,w^{(i)}_{|D_i|}]$, is consists of a sequence of words $w^{(i)}_k$, which denote $k$-th word in $i$-th document. 
\end{definition}

\begin{definition}[\textbf{Hierarchical Discipline Structure}]
The hierarchical discipline structure, expressed as $\gamma$, is a DAG or a tree that consists of disciplines with different levels. On $\gamma$, each node is a discipline, the links between nodes are uniform and directed, pointing from parent nodes to the child nodes, and the child node is a subdiscipline of the parent node. 
Formally, 
the node on this tree is organized in $H$ hierarchical levels $\mathbb{C}=(C_1,C_2,...,C_H)$, where $|\mathbb{C}|$ is the total number of disciplines, and $H$ is the total depth of hierarchical level, $C_i=\{c_1,c_2,...,c_{|C_i|}\}$ is the set of the disciplines and $|C_i|$ is the number of disciplines in the $i$-th hierarchical level.
In order to describe the connection between different disciplines, we introduce $\prec$, a partial order representing the \textit{Belong-to} relationship. $\prec$ is asymmetric, anti-reflexive and transitive\cite{wu2005learning}:

\begin{small} 
 \begin{align} 
   & \bullet \textit{The only one greatest category 'root' is the root of the tree.}  \nonumber\\
   & \bullet \forall c_x \in C_i, c_y \in C_j, c_x \prec c_y  \textit{ then } \space c_y \not\prec c_x.  \nonumber\\
   & \bullet \forall c_x \in C_i, c_x \not\prec c_x. \nonumber\\
   & \bullet \forall c_x \in C_i, c_y \in C_j, c_z \in C_k, \textit{ if } c_x \prec c_y \land c_y\prec c_z \textit{ then } c_x \prec c_z.  \nonumber
   \nonumber
 \end{align}
\end{small} 
\end{definition}

Finally, we define the Hierarchical Discipline Structure $\gamma$ over $\mathbb{C}$ as a partial order set $(\mathbb{C},\prec)$.

\begin{definition}[\textbf{Hierarchical Multi-label Proposal Classification}]
Given a proposal $\mathbb{A}$, we aim to extract its raw data and assign multiple ApplyID codes for it from the hierarchical discipline structure $\gamma$. By treating each discipline in $\gamma$ as a label, we can view this problem as assigning a label set including multiple disciplines for each input proposal, those disciplines in the label set belong to one of the paths in the hierarchy of $\gamma$, where the paths are directional and started from the root node with arbitrary length. 
Formally, the label set is denoted as $L = \{l_1, l_2,..., l_{H_A}\}$, where $l_i\in C_i$ is a disciplines in $\gamma$, $H_A$ is the length of the label set $L$, $\forall l_i,l_j\in L, i > j \to  l_i \prec l_j$.
Moreover, we consider that the expert knowledge $\vartheta$ can be introduced to improve the label assignment. 
Finally, the HMPC problem can be formulated as:

\begin{equation}
\centering
    \Omega(\mathbb{A},\gamma,\vartheta,\Theta) \to L,
\end{equation}
where $\Theta$ is the parameters of model $\Omega$. Figure 1 demonstrates a toy model of the HMPC problem.

\end{definition}

\section{PROPOSED MODEL}
\begin{figure*}[htbp]
\includegraphics[width=\textwidth]{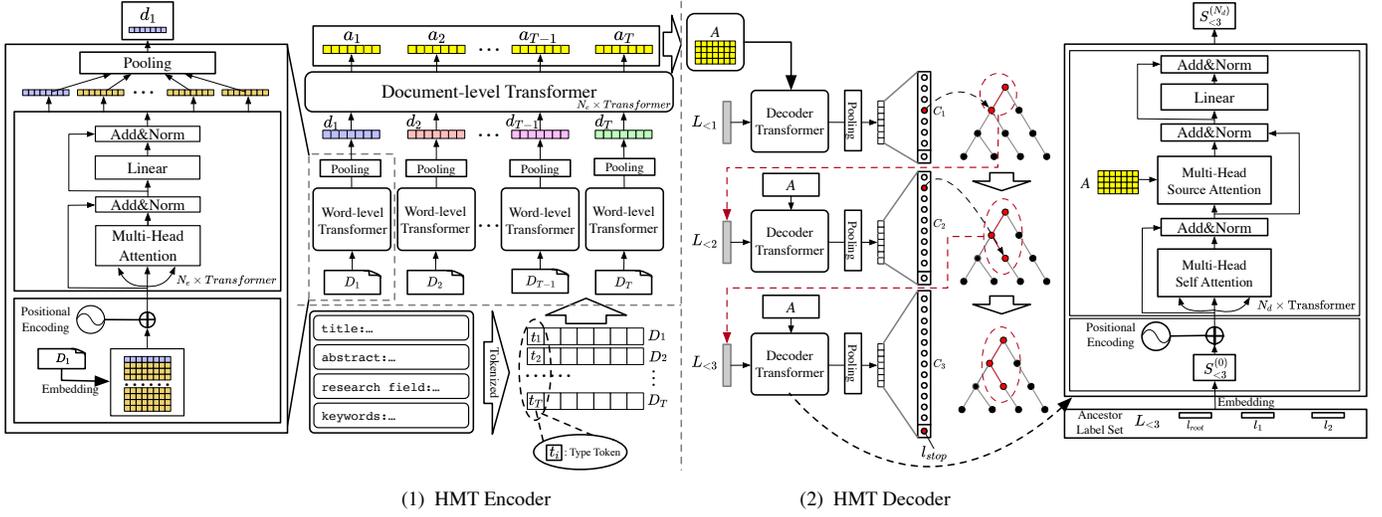} 
\caption{The HMT framework takes a given proposal $\mathbb{A}$ as input and output its label set $L$ on the hierarchical discipline structure. (1) The HMT Encoder. The document set $\{D_1, D_2, ...,D_T\}$ of proposal $\mathbb{A}$ is encoded into the proposal representations $A$. The left side is the detail of the Word-level Transformer in encoding $D_1$ in to the document representation $d_1$. (2) The HMT Decoder. The encoder proposal representations $A$ is input into the Decoder Transformer with the ancestor label set $L_{<k}$ to predict the label in $k$ level. The prediction starts from the root level, until to the 3-th level, the decoder predict the label $l_{stop}$ and stop. The right side is the detail of the Decoder Transformer in the 3-th prediction step.}
\label{fig.models}
\end{figure*}


\subsection{Model Overview}

In our model, the set of documents of the proposal $\mathbb{A}$ serves as the input, and the generated output is a set of labels  $L = \{l_1, l_2,..., l_{H_A}\}$ with length $H_A$, which describes a path from the root to the most fine-grained discipline of the proposal on the hierarchical discipline structure $\gamma$.
As shown in Figure \ref{fig.models}, we first encode the document set of the proposal into vectorized representations. 
Then, we decompose the prediction into an up-to-down generation process from the beginning level to a certain level on the hierarchical discipline structure. 
In each level of prediction, given the proposal representations and the obtained labels from the previous steps, the decoder computes the conditional probability of the labels for the proposal on the current level.
For the first level, we add $l_{root}$, the root node of $\gamma$, as the input label to predict the first label $l_1$. Also, to define the probability over the label length $H_\mathbb{A}$, we add a particular end-of-prediction label $l_{stop}$ to the label set $L$, which enables the iterative predictions of the model to stop on the step when the label $l_{stop}$ is predicted. Thus, the label set $L$ is reset to $L = \{l_{root}, l_1, l_2,..., l_{H_\mathbb{A}}, l_{stop}\}$.
Suppose the $k-1$ ancestors labels in $L$ is $L_{<k} = \{l_{root}, l_1, l_2,..., l_{k-1}\}$, where $L_{<k}\subset L$ and $L_{<1}=\{l_{root}\}$,
the prediction on level $k$ can be formed as: $\Omega(\mathbb{A},L_{<k},\gamma,\Theta) \to l_{k}$.
Moreover, we consider expert knowledge $\vartheta$ given by humans, which are $m$ ground-truth ancestor labels on $L$, formulated as $L_{<m}$.
The prediction would start from the $m$-th level when the expert knowledge is given. Otherwise, when the expert knowledge is not given, it would start from the root of hierarchical discipline structure $\gamma$, i.e., $m=1$. 
Eventually, we can formulate the probability of the assignment of the label set for the proposal as:
\begin{equation}
    P(L|\mathbb{A},\vartheta;\Theta)=\prod_{k=m}^{H_\mathbb{A}+1} P(l_k|\mathbb{A},L_{<k}; \Theta)
    \label{equation_1}
\end{equation}

where $P(L|\mathbb{A},\vartheta;\Theta)$ is the overall probability of the proposal $\mathbb{A}$ belonging to the label set $L$ given the expert knowledge $\vartheta$, $P(l_k|\mathbb{A},L_{<k}; \Theta)$ is the label probability of $\mathbb{A}$ in level $k$, given the previous ancestor labels before the level $k$.
In training, given all the ground truth labels, our goal is to maximize the Equation 2.

In this way, our model can be regarded as an encoder-decoder framework. Similar to the translation model, the encoder extracts textual information from the raw data of proposals, and the decoder generates the labels of proposals on the hierarchical discipline structure.
For model implementation, we introduce the Transformer model proposed in \cite{vaswani2017attention}, which is a state-of-the-art module and has demonstrated very competitive ability in combination with encoder-decoder based models\cite{devlin2018bert, vaswani2017attention}. In the following, we will first introduce the architecture of the Transformer. Then, we will introduce the encoder and decoder design in detail.

\subsection{Transformer Details}
A Transformer model usually has multiple layers. A layer of transformer model (i.e, a Transformer block) consists of a \textit{Multi-Head Self-Attention Mechanism}, a \textit{Residual Connections and Layer Normalization Layer}, a \textit{Feed Forward Layer}, and a \textit{Residual Connections and  Normalization Layer}, which can be written as:

\begin{equation}
\label{eq:3}
    Z = \sigma(X^{(l)}+MultiHead(X^{(l)},X^{(l)},X^{(l)}))
\end{equation}

\begin{equation}
\label{eq:4}
    X^{(l+1)} =  \sigma(Z+\mathcal{F}(Z))
\end{equation}

where $X^{(l)} = [x_1^{(l)}, x_2^{(l)}, ..., x_s^{(l)}]$ is the input sequence of $l$-th layer of Transformer, $s$ is the length of input sequence, $ x_i^{(l)}\in \mathbb{R}^h$ and $h$ in the dimension. $\sigma(\cdot)$ is layer normalization, $\mathcal{F}(\cdot)$ denotes a two-layer feed-forward network with $ReLU$ activation function, and $MultiHead(\cdot)$ denotes the multi-head attention mechanism, which is
calculated as follows:

\begin{equation}
\label{eq5}
MultiHead(Q, K, V) = Concat (head_1, . . . , head_h)W^O
\end{equation}

\begin{equation}
head_i = Attention(Q W_i^Q,K W_i^K,V W_i^V)
\end{equation}

\begin{equation}
\label{eq7}
Attention (Q,K,V) = softmax(\frac{QK^T}{\sqrt{d}})V
\end{equation}


where $W_i^Q, W_i^K, W_i^V\in \mathbb{R}^{h\times h}$ are weight matrices and $d$ is the number of attention heads. After the attention calculation, the $h$ outputs are concatenated and
transformed using a output weight matrix $W^O\in \mathbb{R}^{dh\times h}$.

In summary, given the input token sequence $X = [x_1, x_2, ..., x_s]$, the Transformer first input it into a embedding layer to obtain the initial embeddings of this sequence $X^{(0)} = [x_1^{(0)}, x_2^{(0)}, ..., x_s^{(0)}]$. Then, after the propagation in Equation \ref{eq:3} and \ref{eq:4} on a $N$-layers Transformer, formulized as $X^{(N)} = Transformer(X)$, we can obtain the final output embeddings of this sequence $X^{(N)} = [x_1^{(N)}, x_2^{(N)}, ..., x_s^{(N)}]$, where each embedding has contained the context information in this sequence.  The main hyper-parameters of a Transformer are the number of layers (i.e., Transformer blocks), the number of self-attention heads, and the maximum length of inputs. 

\subsection{HMT Encoder}
To extract the semantic information in a given Proposal $\mathbb{A}$, a simple solutions is to concatenate all its words into a flat text sequence as the input of Transformers. 
However, the proposal may have abundant text information with different types. 
On the one hand, the input sequence would be extremely long, which would be computationally intensive for the encoding of the Transformer according to Equation \ref{eq5}-\ref{eq7}. 
On the other hand, the simple concatenation ignores the uniqueness between different text types, which would undermine the model effectiveness. 
Thus, we propose a hierarchical textual encoder for proposal feature extracting. Figure \ref{fig.models}(1) shows the detail of the HMT Encoder framework. The HMT Encoder consists of a word-level Transformer encoder and a document-level Transformer encoder.
The raw proposal data $\mathbb{A}$ is split into multiple documents $\{D_1, D_2, D_3,...,D_{T}\}$ in terms of document types, which are passed through the word-level Transformer encoder separately to get the type-specific document representations. 
Then, the document-level Transformer encoder is proposed to extract the document features and generate the representation matrix $A$ of the input proposal $\mathbb{A}$.


\subsubsection{Word-level Transformer}
In the first stage of HMT Encoder, we aim to use a multi-layer Transformer to embed the text in $D_i$ into the document representation $d_i \in \mathbb{R}^h$.
We first add a type-specific token $t_i$ at the beginning of each document $D_i$ to mark its type. 
Thus, for each document $D_i$ with token sequence $[t_i, w^{(i)}_1,w^{(i)}_2,...,w^{(i)}_{|D_i|}]$,
the document representation vector $d_i$ can be obtained by:
\begin{equation}
\begin{aligned}
    d_i &= Pooling([e^{(i)}_t, e^{(i)}_1, ..., e^{(i)}_{|D_i|}])\\
    &= Pooling({Transformer_{e_1}([t_i,w^{(i)}_1,..., w^{(i)}_{|D_i|})}]) 
\end{aligned}
\end{equation}
where $Transformer_{e_1}(\cdot)$ is a $N_e$-layers Transformer to extract the word-level text feature. For each token ${w}^{i}_j \in D_i$ and the type token $t_i$ in the sequence, its input vector is constructed by summing its embedding with its corresponding positional embedding in documents. The word embedding is initialized by pre-trained $h$-dimensional Word2Vec\cite{mikolov2013distributed} model and the type embedding is random initialized. 
$[e^{(i)}_t, e^{(i)}_1, ..., e^{(i)}_{|D_i|}]$ is the list of output embeddings.  The $Pooling(\cdot)$ is the pooling operation on the outputs to obtain $d_i$. In our paper, we define this pooling operation as directly taking the head vector $e^{(i)}_t$ as the document representation $d_i$, as $e^{(i)}_t$ hold by type $t_i$ indicates the type of $D_i$ that incorporates the textual information in the context of document. 


\subsubsection{Document-level Transformer}
The document-level in the second stage of HMT Encoder aims to encode the document-level representations for proposals. Given $\mathbb{A}$'s document set $\{D_1, D_2, ... ,D_{T}\}$, we can obtain all the documents' representations set $\{d_1, d_2, ... ,d_{T}\}$ from the above word-level Transformer. Then, we put the documents' embeddings to the document-level Transformer: 

\begin{equation}
\begin{aligned}
    A & = [a_1, a_2, ..., a_T] \\
    &= {Transformer}_{e_2}(\{d_1, d_2, ... ,d_{T}\}), 
\end{aligned}
\end{equation}
where ${Transformer}_{e_2}(\cdot)$ is a $N_e$-layers Transformer to extract the document-level features for proposals, $A = [a_1, a_2, ..., a_T]$ is the Transformer outputs, which can be reviewed as $T$-views of representations for proposals. Contributed by the self-attention mechanism in Transformer, each document representation in the inputs can collect information from the context documents with different attention weights, so the corresponding output contains the information of the whole document set, which can be regarded as a facet of representation for proposals.

Be noted that the encoder is a textual feature extractor essentially. 
We design the HMT Encoder in the consideration of both efficiency and effectiveness, which mainly has the following advantages:
(1) Comparing with other RNN-like models as text encoder, Transformer is more efficient to learn the long-range dependencies between each word. 
Specifically, with the help of two-level Transformer encoder, we can obtain the associated proposal representations with long-range dependencies in both the word-level and the document-level. 
(2) Also, RNN-like models extract text information via sequentially propagation on sequence, which need serial operation of the time complexity of sequence length. While self-attention layers in Transformer connect all positions with constant number of sequentially executed operation. Therefore, Transformer can encoder text in paraller.
(3) The types of different textual features of proposals are taken into consider via adding the type-token in proposals' documents, which helps Transformer to extract type-specific outputs of documents. The design of two-level Transformers reduces the number of parameters of the encoder, which greatly improves the training efficiency of the entire model.


\subsection{HMT Decoder}
After obtaining the representations $A$ of the proposal, we propose the HMT Decoder to generate the label set of the proposal.
According to Equation 2, we hope that the decoder can generate labels in a top-down fashion from level $m$ to level $H_A$ on the hierarchical discipline structure $\gamma$, with considering both the proposal information and the obtained ancestor labels. Thus, the decoding process is decomposed into multi-steps, in each step $k$, we input the proposal representations $A$ and the ancestor labels $L_{<k}$ to generate the label probability $P(l_k|\mathbb{A},L_{<k}; \Theta)$ of proposal $\mathbb{A}$ on the $k$-th level. Figure \ref{fig.models}(2) demonstrate the detail of the HMT Decoder framework. The decoder is also based on Transformer, which is consisted with multiple layers, its architecture is similar with the encoder's Transformer, but it introduce a source attention layer in addition to the self-attention layer. The decoder Transformer aims to extract the ancestor label information of $L_{<k}$ and learn the dependency of labels on the hierarchical discipline structure. Also, it integrates the proposal representations and ancestor label information via the source attention layer. After several layer's propagation, the decoder is connected to a multi-layer perceptron with softmax to generate the label probability on the current hierarchical level.


Formally, we first random initialize the embeddings of labels, we sum the initial embeddings of the ancestor label set $L_{<k}=\{l_{root},l_{1},l_{2},...,l_{k-1}\}$ with corresponding positional embeddings as the Transformer input $S_{<k}^{(0)} =[s^{(0)}_{root},s^{(0)}_{1},s^{(0)}_{2},...,s^{(0)}_{k-1}]$. Given $S_{<k}^{(0)}$ and proposal representations $A$, the propagation of the $l$-th layer of the decoder Transformer is defined as:

\begin{equation}
\label{M}
    \widehat{S_{<k}^{(l)}} = \sigma(S_{<k}^{(l)}+ MultiHead(S_{<k}^{(l)},S_{<k}^{(l)},S_{<k}^{(l)}))
\end{equation}


\begin{equation}
\label{Z}
    Z = \sigma(\widehat{S_{<k}^{(l)}} + MultiHead(\widehat{S_{<k}^{(l)}} , A, A))
\end{equation}

\begin{equation}
\label{L}
    S_{<k}^{(l+1)} =  \sigma(Z+\mathcal{F}(Z))
\end{equation}

The propagation can be divided into two stages. In Equation \ref{M}, the input label is feeded into a multi-head self-attention layer to capture the dependency between labels in the label set, and the output $\widehat{S_{<k}^{(l)}}\in \mathbb{R}^{k\times h}$ preserves the ancestor labels information, which is utilized in the second stage. In Equation \ref{Z} and \ref{L}, $\widehat{S_{<k}^{(l)}}$ is incorporated with the proposal representation matrix $A$ via a source attention layer, where $\widehat{S_{<k}^{(l)}}$ is the query, $A$ is the key and value, which aims to extract information from the proposal corresponding to the predicted labels. The output of this layer is the hidden embeddings $S_{<k}^{(l+1)} =[s^{(l+1)}_{root},s^{(l+1)}_{1},s^{(l+1)}_{2},...,s^{(l+1)}_{k-1}]$, where each $s^{(l+1)}_i$ can be seem as a representation of the proposal which is correlated with the label $l_i$. 
Finally, We stack $N_d$ layers of above Transformer as the HMT Decoder and obtain the final output in last layer:

\begin{equation}
\begin{aligned}
    S_{<k}^{(N_d)} & = [s^{(N_d)}_{root},s^{(N_d)}_1,s^{(N_d)}_2,...,s^{(N_d)}_{k-1}] \\
    &= {Transformer}_d(\{l_{root},l_{1},l_{2},...,l_{k-1}\}, A)
\end{aligned}
\end{equation}

where $S_{<k}^{(N_d)}\in \mathbb{R}^{k\times h}$ has integrated the proposal information with $k$ previous ancestor labels of the proposal, which can be seem as $k$ proposal representations corresponding to the $k$ levels. Then, we feed $S_{<k}^{(N_d)}$ into a pooling function, a multi-layer perceptron layer and a softmax layer to generate the label probability on the $k$-th hierarchical level:

\begin{equation}
 {\hat{y}}_k= Softmax({\mathcal{F}}_{k}(Pooling(S_{<k}^{(N_d)})))
\end{equation}

The $\mathcal{F}_{k}(\cdot)$ denotes a level-specific feed-forward network with ReLU activation function to project the input to a $|C_k|+1$ length vector. After the $Softmax(\cdot)$, the final output $\hat{y}_k$ is the probability of $k$-th level's labels, in validation, the label corresponding to the index of the largest value in the $\hat{y}_k$ will be the predicted label. In our paper, we set this pooling operation as directly taking the last vector of $S_{<k}^{(N_d)}$, as its level is the most is the closest to the 
level $k$ where the label needs to be predicted.
Noted that we require that the model is able to define a distribution over labels of all possible lengths, we add a special end-of-prediction label $l_{stop}$ to all levels, and set the last element of the probability in each level is the probability of label $l_{stop}$. As shown in left side of Figure \ref{fig.models}(b), in the last step, the $l_{stop}$ is the prediction result, the prediction process should end in the level-$n$ and output the current predicted labels as the final result. Thus, suppose index of the groud truth label $l_k$ in $k$-th level is $i$, the label probability can be obtained by $P(l_k|\mathbb{A},L_{<k}; \Theta) = \hat{y}_{k,i}$. 
Finally, after the prediction from level $m$ to the level $H_A$,  the standard training objective for each proposal $\mathbb{A}$ is to maximize the log-likelihood of the overall probability:

\begin{equation}
    \Theta = \mathop{\arg\max}_{\Theta}\sum_{k=m}^{H_\mathbb{A}+1} \log P(l_k|\mathbb{A},L_{<k}; \Theta)
\end{equation}

To our best knowledge, we are the first to utilize the Transformer model to perform the hierarchical multi-label classification.
The main advantages of our designed decoder can be summarized as follows: 
(1) The Transform model has been proved to be superior to align the semantic and syntactic information from textual sequences. We believe our decoder can learn the dependencies between labels on the hierarchical discipline structure.
(2) As the ancestor labels are easier to predict than the low-level labels. The decoder is able to use the previews predicted ancestor labels to help the prediction in low level. Also, it allows experts to assign the ancestor labels. 
(3) By introducing the end-of-prediction label $l_{stop}$, the decoder gained the capacity of predicting the label length of proposal, which means the decoder can automatically learn the proper granularity of discipline for proposals.

\section{Experiments}
In this section, we evaluate our HMT model on the hierarchical multi-label proposal classification problem. We main focus on answer the following questions about our model: (1) Can HMT achieve very accurate prediction results on real-world proposal datasets? (2) Does HMT have the ability to use expert knowledge to improve the prediction performance? (3) Can HMT predict the label length right? i.e., assign proposal disciplines in a proper granularity? (4) Can HMT learn the label dependencies on the hierarchical discipline structure? Besides, we conduct ablation study, hyperparameter sensitivity, and model convergence experiments to explore the analyze the underlying mechanism of HMT. The HMT code and corresponding data is available online.\footnote{the code and desensitized data is shared on here: \href{https://www.dropbox.com/sh/007s3yo1eldh359/AAC-yE_z6umInEI5rKMv0T8ca?dl=0}{[link]}}.

\subsection{Dataset}
In our experiments, we collected 244,800 research proposals of 3499 ApplyIDs from the 2019 proposals database in the National Science Foundation China.
The detail of the association between ApplyID prefix code and Department are listed in the Table \ref{tab.prefix_stat}, which can observe that the ApplyID codes in the ApplyID system are organized into four levels structure. 136,477 proposals' most fine-grained label is on level-4, 103,768 on Level-3's, and 4,555 on Level-2's. The Level-1 ApplyID code refers to the main discipline, so no proposal is directly located on the first level's ApplyID code. 
This database is named \textit{NSFC-19}. Like other HMC model's setting, in our experiments, we used 80\% proposals from \textit{NSFC-19} as train data, the 10\% as validation data and 10\% test data. The text data of all the proposals have been desensitized and deidentified. As we described before, the proposal data are composed of four different parts: 1) title, 2) keywords, 3) research fields, 4) abstract. The title, research fields, and abstract is in a form of long text. 


\begin{table}[htbp]
\centering
\caption{ApplyID Prefix Code and its Associate Department}
\scalebox{0.90}{
\begin{tabular}{|c|c|c|c|c|c|}
\hline
Prefix & Discipline Name                                      & Total & $|C_2|$ & $|C_3|$ & $|C_4|$ \\ \hline
A                & Mathematical Sciences              & 318 & 6 & 57 & 255      \\ \hline
B                & Chemical Sciences                   & 396 & 8 & 59 & 329  \\ \hline
C                & Life Sciences                      & 827 & 21 & 189 & 617  \\ \hline
D                & Earth Sciences                     & 167 & 7 & 95 & 65 \\ \hline
E                & Engineering and Materials Sciences & 434 & 9 & 116 & 309 \\ \hline
F                & Information Sciences               & 793 & 7 & 73 & 713  \\ \hline
G                & Management Sciences                & 107 & 4 & 57 & 46 \\ \hline
H                & Medicine Sciences                      & 457 & 30 & 427 & 0 \\ \hline
- & Total Discipline & 3499 & 92 & 1073 & 2334 \\ \hline
\end{tabular}}
\label{tab.prefix_stat}
\end{table}

%

\begin{table*}
\centering
\caption{Performance of the different model}
\resizebox{\textwidth}{!}{%
\begin{tabular}{c|cc|cc|cc|cc|cc}
\hline
\multirow{2}{*}{Baseline} & \multicolumn{2}{c|}{Level 1}                    & \multicolumn{2}{c|}{Level 2}                    & \multicolumn{2}{c|}{Level 3}                    & \multicolumn{2}{c|}{Level 4}                    & \multicolumn{2}{c}{Overall}                               \\ \cline{2-11} 
                          & \multicolumn{1}{c|}{Micro-F1} & Macro-F1       & \multicolumn{1}{c|}{Micro-F1} & Macro-F1       & \multicolumn{1}{c|}{Micro-F1} & Macro-F1       & \multicolumn{1}{c|}{Micro-F1} & Macro-F1       & \multicolumn{1}{c|}{Micro-F1} & Macro-F1                   \\ \hline
HMCN-F & 0.757 & 0.726 & 0.472 & 0.383 & 0.270 & 0.152 & 0.488 & 0.098 & 0.479 & 0.126
\\
HMCN-R &  0.761                         & 0.728          & 0.476                         & 0.391          & 0.277                         & 0.173          & 0.487                         & 0.110          & 0.500                         & 0.141
\\
HARNN                     & 0.909                         & 0.890          & 0.845                         & 0.821          & 0.792                         & 0.721          & 0.860                         & 0.711          & 0.852                         & 0.718                     \\
\hline
$\text{HMCN-F}_t$                    & 0.962                         & 0.952          & 0.916                         & 0.891          & 0.832                         & 0.738          & 0.865                         & 0.674          & 0.894                         & 0.703                      \\
$\text{HMCN-R}_t$                   & 0.960                         & 0.951          & 0.917                         & 0.901          & 0.851                         & 0.772          & 0.884                         & 0.753          & 0.903                         & 0.764                      \\ 
$\text{HARNN}_t$ & 0.950 & 0.940 & 0.915 & 0.901 & 0.857 & 0.783 & 0.889 & 0.752 & 0.903 & 0.767\\ 

\hline
HMT-FLAT & 0.983 & 0.979 & 0.689 & 0.667 & 0.736 & 0.698 & 0.569 & 0.286 & 0.744 & 0.437
\\
HMT-CONCAT & 0.977 & 0.971 & 0.952 & 0.938 & 0.896 & 0.843 & 0.884 & 0.802 & 0.930 & 0.820
\\
\hline
HMT                     & \textbf{0.983}                         & \textbf{0.979}          & \textbf{0.962}                         & \textbf{0.952} & \textbf{0.910}                & \textbf{0.862} & \textbf{0.900}                & \textbf{0.824}          & \textbf{0.939}                         & \textbf{0.841}             \\ \hline  
\end{tabular}%
}
\label{table:exp_baseline}
\end{table*}
\subsection{Baseline Methods}
To comprehensively evaluate the performances of our proposed method under the HMPC task, we compare it with the three state-of-the-art Hierarchical Multi-label Classification(HMC) models, the HMCN-F, HMCN-R and HARNN. Those baseline models utilize the hierarchical information between each level and can perform on texts. The baseline models we used in the experiments are listed below:\\
\textbf{Hierarchical Multi-label Classification model:}
    We compared HMT with three related hierarchical multi-label classification (HMC) models and its variant with HMT Encoder.
    \begin{itemize}
        \item \textbf{HMCN-F and HMCN-R\cite{wehrmann2018hierarchical}:} The HMCN-F is an HMC model that uses a feed-forward network and hybrid prediction layer. The hybrid prediction layer predicts not just category labels of each level and but also the overall category label. HMCN-R is an RNN-like variant of HMCN-F. The original HMCN method haven't provided encoder method, so we use \textit{Doc2Vec}\cite{doc2vec} as the input feature of HMCN. 
        \item \textbf{HARNN\cite{huang2019hierarchical}} The HARNN consists of a LSTM\cite{lstm} based encoder and a hierarchical attention-based memory unit as classifier. The classifier unit preserves the last prediction result to generate next prediction. The model will make prediction on global hierarchy structure in the last prediction. 
        \item \textbf{$\text{HMCN-F}_t$, $\text{HMCN-R}_t$ and $\text{HARNN}_t$:} These baseline models are the variant of HMCN-F, HMCN-R, and HARNN, which the HMT Encoder as their encoder. 
    \end{itemize}
\textbf{HMT Variants:}
    We post two variants of HMT as baselines:
    \begin{itemize}
        \item \textbf{HMT-FLAT:} 
        a variant of HMT, which ignores the hierarchical strucutre of labels and treats the problem as a traditional multi-label classification. It replaces the decoder into a flat classifier. 
        \item \textbf{HMT-CONCAT:} a variant of HMT, which ignores the different document type in proposal. It concatenate all the text into one sentence and use a Transformer as encoder.
    \end{itemize}

\subsection{Experiment Setting.}
We set the HMT Encoder layer number $N_e$ to 8, the dimension size $h$ to 64, and the multi-head number to 8. The HMT Decoder layer number $N_d$ is set to 1, and the multi-head number is set to 8. We statistic the length of the long text in every document, for each long text form documents, we set the max sequence length as 50 in our model. For training HMT, we use Word2Vec\cite{mikolov2013distributed} model with Gensim tools\cite{gensim} with a dimension ($h$) 64 to generate the word embedding for each Chinese characters in a document on all text in \textit{NSFC-19}. For the detail of HMT training, we use Adam optimizer\cite{kingma2014adam} with learning rate of $1\times 10^{-3}$, and set the mini-batch as 512, adam weight decay as $1 \times 10^{-5}$. The dropout rate is setting to $0.2$ to prevent overfitting. The warm up step is setting as 1000. In the following experiments, all methods is implemented by PyTorch\cite{paszke2019pytorch} and all experiments are conducted on a Linux server with a AMD EPYC 7742 CPU and eight NVIDIA A100 GPUs. For fair comparison, all hyperparameters in these baselines are tuned to achieve the best performance. All the experiments are repeated many times to make sure the results can reflect the performances of methods.
\subsection{Experiment Results}

\noindent\textit{\textbf{1. Performance Comparison:}}

The first experiment is the overall performance comparison and the performance comparison on each level between our method and baseline methods. We compared HMT with all the baseline methods on the HMPC task using the \textit{NSFC} dataset. The results are shown in Table \ref{table:exp_baseline}.

\textbf{Overall performance:}
We first introduced the performances of our method and baseline methods on the overall level. In this comparison, we organized each prediction result flatly, then we used ${\textit{Macro-F}_1}$ and ${\textit{Micro-F}_1}$ as evaluation metrics, which are widely used in hierarchical multi-label classification problem \cite{metric1}\cite{metric2}\cite{huang2019hierarchical}. From the overall results, we observed that:

1) Our method HMT achieves the best performance on the \textit{NSFC-19} dataset in overall evaluation metrics, which indicates that HMT is better in modeling both the complex textual documents and hierarchical category structure. 
    
2) Our method HMT outperforms HMT-FLAT, $\text{HARNN}_t$, $\text{HMCN-F}_t$, and $\text{HMCN-R}_t$ on overall evaluation metrics with the same encoder of $\text{HMT}$ in the NSFC-19 dataset, which proves that the decoder of $\text{HMT}$ can better utilize the dependency of labels among all levels in the hierarchical discipline structure to make prediction than other baseline methods.
    
3) The $\text{HARNN}_t$, $\text{HMCN-F}_t$, and $\text{HMCN-R}_t$ perform better than the origin models on overall evaluation metrics. The reason is that, compared with the \textit{LSTM} based encoder in $\text{HARNN}$ or the documents representation pretrained by \textit{Doc2Vec} in $\text{HMCN-F}$ and $\text{HMCN-R}$, the encoder of $\text{HMT}$ can better extract the feature from the proposals.  
    
4) HMT beats HMT-FLAT by leveraging the hierarchical discipline structure information, which indicates that the consideration of the hierarchical discipline structure in HMT is beneficial to the HMPC task.
    
5) The HMT performs better than HMT-CONCAT by utilizing the type information inside the proposal with the type-token, which can prove that using the information of the document type is beneficial to model prediction results. Also, compared with the encoder of HMT-CONCAT that concatenates all text sequences' representation together, the HMT Encoder only passes the type-token's vector to the decoder, thus can reduce the model's calculation cost.

\textbf{Accuracy on each level:}
In the HMPC task, annotating categories of each level correctly is critical. Thus we conduct an experiment to compare HMT with other baseline models on each level of the hierarchical discipline structure separately. Table \ref{table:exp_baseline} shows that HMT outperforms the baseline methods and variants on  most category levels. 
Moreover,  the accuracy of HMT and its variants tend to decrease when the depth increases, but HMT shows better performance than other variants. 
Indeed, when the hierarchical level depth increase, the number of categories labels  increase rapidly(e.g., the \textit{NSFC-19} dataset has 92 and 1073 category labels on the level-2 and level-3, respectively, as shown in Table \ref{tab.prefix_stat}), which significantly influences the model performance. 
HMT is able to model the document type and the hierarchy information  more effectively than other variants. 
In addition, the accuracy of HMCN-F, HMCN-R, HARNN, and their variants tend to decrease at the first three levels and increase at the level-4, showing their designed models put more focus on prediction of the last level.



\noindent\textit{\textbf{2. the Utilization of Expert Knowledge}}

Figure \ref{Fig.label_1} demonstrates experts can provide partial labels to enhance proposal classification. In real-world HMPC tasks, it is critical to utilize this partially available label information provided by domain experts to generate more accurate and fine-grained labels. Thus, in Experiment 2, we evaluated the HMT's ability to utilize the partial information provided by experts. 
We first initialize the three kinds of partial label set $L_{<i}$ by the ground-truth labels, where $i\in \{1,2,3\}$, which represents the coarse-grained ancestor labels provided by experts. 
Then, HMT will generate the rest labels step by step. We evaluate the overall performances  and each level's $\textit{Micro-F}_1$ and $\textit{Macro-F}_1$. 

\begin{figure}
\centering
\includegraphics[width=0.48\textwidth]{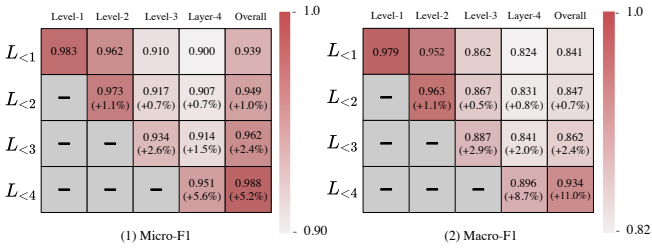} 
\caption{The Results on Different Given Expert Knowledge}
\label{Fig.hotmap}
\end{figure}

The heatmap of the results is listed in Figure \ref{Fig.hotmap}, where every row represents the performance of HMT on different given partial labels, respectively. The columns represent the performance on different levels and overall performance. The color in Figure \ref{Fig.hotmap} changes from white to red indicating the accuracy increases. From the figures, we can observe that: (1) By using the partial information, HMT can generate better categories labels for rest levels. This phenomenon explains that HMT has the ability to use partially available coarse-grained information to generate the rest fine-grained labels. (2) Furthermore, with the information provided, the overall and all levels' $\textit{Micro-F}_1$ and $\textit{Macro-F}_1$ increase, which means the partial information can help HMT make the rest prediction more precisely. This is because HMT will use the previous prediction results when making next level label prediction, and the accurate prior information provided by coarse-grained information can reduce errors caused by incorrectly previous prior prediction.\\
\noindent\textit{\textbf{3. Path Length Sensitivity Study}}

\begin{figure}
\includegraphics[width=0.45\textwidth]{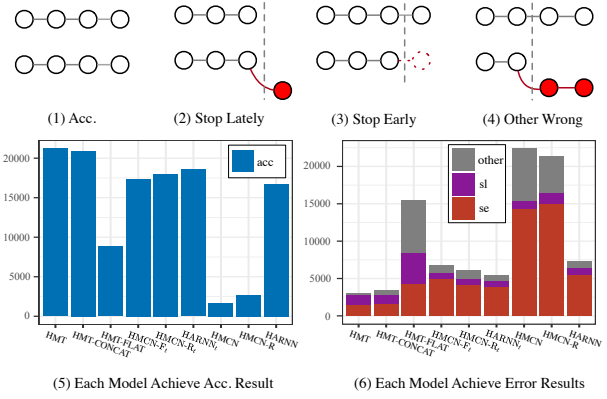} 
\caption{The Result of Path Length Sensitive Study}
\label{Fig.gra_result}
\end{figure}

In this experiment, we aim to evaluate the HMT's and other baselines' ability to give the proper granularity of prediction. 
As we mentioned before, a proposal will be assigned to a path of labels, the longer the path is, the finer-grained label will be assigned. We use the length of the predicted label path to indicate the granularity of the prediction. By that, Figure \ref{Fig.gra_result}(1-4) demonstrate four different types of prediction results. The Accuracy (\textit{acc}) state mean model gives the right result and proper granularity for the input proposal. The Stop Lately (\textit{sl}) and Stop Early (\textit{se}) results stand for model give a longer or shorter prediction result but with a correct prediction previously. This means that the model tends to classify the proposal into a finer-grained or coarser-grained hierarchy label. And Other Wrong (\textit{other}) means the model makes a wrong prediction during the whole progress. The more \textit{sl}, \textit{se}, and \textit{other} occurred, the less path length sensitive the model have. 

We compare HMT with all baselines. From the experiment result listed in Figure \ref{Fig.gra_result}(5-6), we can observe that: (1) HMT gives the most \textit{acc} results, which means that compared to other baseline models, HMT can give the most proper granularity and most correct prediction. (2) The $\text{HARNN}$, $\text{HMCN-F}$, and $\text{HMCN-R}$ and their variants tend to be predict shorter label sets, which cause more \textit{se} error. In contrast, HMT, HMT-CONCAT, and HMT-FLAT are more balanced in the probability of occurrence of the two errors. (3) Comparing to other baselines, there are fewer \textit{sl}, and \textit{se} errors occurred in HMT, proves that HMT Decoder has the best length sensitivity. \\
\noindent\textit{\textbf{4. Hierarchy Dependency Sensitivity Study}}

\begin{figure}
\centering
\includegraphics[width=0.48\textwidth]{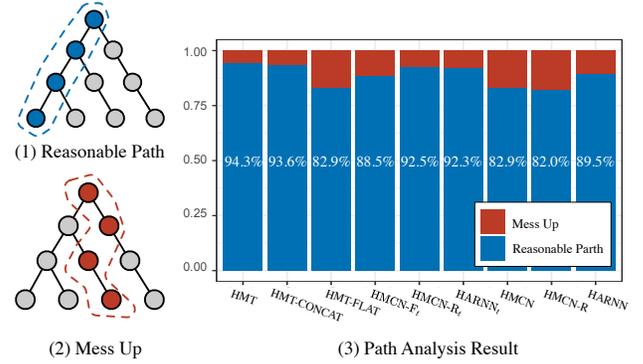} 
\caption{The Result of Hierarchy Dependency Sensitivity Study}
\label{Fig.path.all}
\end{figure}

In this experiment, we evaluate the sensitivity to the hierarchical discipline structure of the model. 
Figure \ref{Fig.path.all}(1),(2) shows two types of prediction result, the \textit{Reasonable Path} is a series of labels that obey the hierarchical structure, i.e., the labels compose a valid path on the tree, and vise versa is the \textit{Mess Up} result. With more the \textit{Reasonable Path} generate, indicate that the model is more efficient to capture the hierarchical dependency information between each level, in another word, the model is more sensitive to the hierarchy dependency.

We evaluate HMT with all baselines. From the result listed in Figure \ref{Fig.path.all}(3), we can observe that: 
(1) The HMT beats other baseline models by generating $94.3\%$ \textit{Reasonable Path}, which means HMT has the best sensitivity to hierarchy dependency. This proves that our model can capture the dependency between each label in the hierarchical discipline structure.
(2) The HMT and HMT-CONCAT that use HMT Decoder outperform HMT-FLAT. The result proves that the HMT Decoder can capture the hierarchy information more efficiently. 
(3) The \textit{Reasonable Path} appear more frequently in $\text{HARNN}_t$, $\text{HMCN-F}_t$ and $\text{HMCN-R}_t$ than their original models. This indicates that HMT Encoder can provide better representation of the proposal for the decoder.

\noindent\textit{\textbf{5. Hyperparameter Sensitivity Study}}

\begin{figure}
\centering
\includegraphics[width=0.49\textwidth]{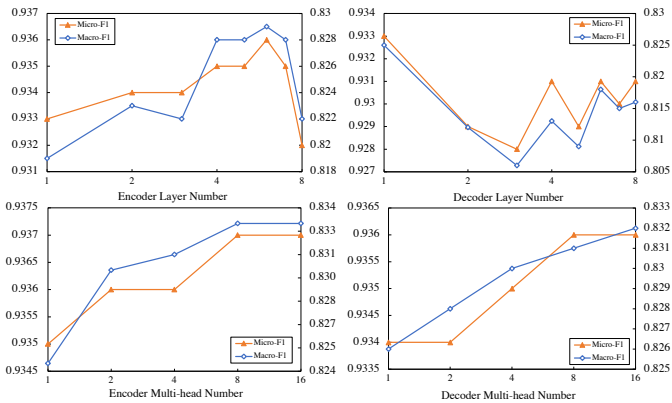} 
\caption{Hyperparameters Sensitivity Study}
\label{Fig.hyper}
\end{figure}

We also conduct experiments to evaluate the effect of four key hyperparameters in our model, i.e, the number of layers the number of self-attention heads in the HMT Encoder and HMT Decoder. We investigate the sensitivity of these two parameters in the encoder and decoder and report the results in Figure \ref{Fig.hyper}, we can observe that: 
(1) When the layers of Transformer in encoder varies from 1 to 8, the performance of HMT increases at first then decreases slowly. That is because with more parameters introduced, the model is likely to over-fitting to impact the classification results.
(2) When the layers of Transformer in the decoder vary from 1 to 8, the performance of HMT decreases slowly. That is because labels contains less information than text, which makes it easier to over-fitting than the encoder. 
(3) The more number of attention heads will generally improve the performance of HMT, while with the further increase of attention heads, the improvement becomes slight. Meanwhile, we also find that more attention heads can make the model more stable.

\noindent\textit{\textbf{6. Model Convergence Experiment}}

\begin{figure}
\centering
\includegraphics[width=0.48\textwidth]{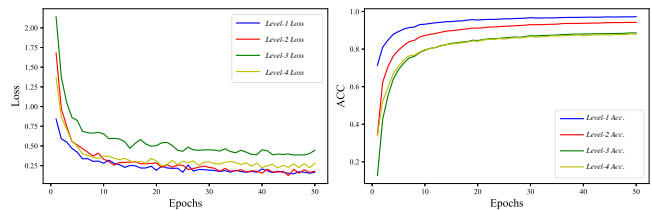} 
\caption{Loss and Accuracy Curves}
\label{Fig.loss_acc}
\end{figure}

To evaluate the training process of our model. We visualize the curves of level-wise loss and accuracy on the training data. The curves of both loss and accuracy show in Figure \ref{Fig.loss_acc}. From the figures, we can observe that the accuracy on the ancestor level yield a litter more faster convergence than the lower level in the beginning, that is because the prediction results on low level is depends on the prediction of ancestor level, where the prediction error will be amplified with the generating across levels. Then, the loss curve of classification in four levels shows the training is stable.

\section{Related Work}
The methods for the HMC problem are designed to cope with the classes, which are organized by asymmetric, anti-reflexive, and transitive relations. Some traditional methods are referred to as 'the direct approach'\cite{burred2003hierarchical} or 'global classifier'\cite{xiao2007hierarchical} use the flat classification method without considering the hierarchy information. Local approach methods\cite{10.1007/978-3-540-85557-6_4} follow the top-down paradigm but ignore the dependency between each level, and error propagate from the upper levels to subsequent ones. Other local approaches have been proposed in the literature, like Local classifier per Node(LCN) or Local classifier per parent node(LCPN). Beside that, several studies on global approaches propose to directly incorporate the hierarchical classification scheme in the learning process, both in standard machine learning\cite{2004Integrating} and in the deep learning\cite{2018Large}\cite{piroi2011clef}.

Currently, the methods for hierarchical multi-label classification mostly focus on design of loss functions or neural network architecture\cite{cerri2015hierarchical}\cite{wehrmann2018hierarchical}\cite{gargiulo2019deep}\cite{huang2019hierarchical}.Giunchiglia et al. leverage standard neural network approaches for multi-label classification problems and then exploit the hierarchy constraint in order to produce coherent predictions and improve performance\cite{giunchiglia2021multi}. 
Other works focus on using other deep learning method. Zhang et al. propose a document categorization method\cite{Zhang2021} with hierarchical structure under weak supervision. The work\cite{tang2020multi} design a attention-based graph convolution network to category the patent to IPC codes. The work in \cite{LaGrassa2021} use the convolution neural network as an encoder and explore the HMC problem in image classification. Zhang et al. propose HiMeCat\cite{Zhang2021} an embedding-based generative framework with a joint representation learning module to categorize documents into a given label hierarchy under weak supervision. The works in \cite{nakano2020active} propose a active learning approach for HMC problem. Aly et al. propose a capsule network based method\cite{Aly2019} for HMC problem. Also there are some work focus on co-embedding the classes and entity into vector space for preserve the hierarchy structure. HyperIM\cite{chen2020hyperbolic} use the hyperbolic space\cite{hamann2018tree} to embeded the tree-like hierarchical structures and illustrate the label dependency in a hierarchical multi-label text classification problem. TAXOGAN\cite{Yang2020} embedding the network nodes and hierarchical labels together, which focus on taxonomy modeling. 

Besides these, several works model the HMC problem as a sequence-to-sequence\cite{sutskever2014sequence} task. For example, Li et al. proposes a sequence to sequence model for hierarchical classification\cite{li2018unconstrained} while the model will also generate the hierarchical structure during the training. The latest related work is proposed by Risch et al.\cite{risch2020hierarchical}. They use word embedding with feed-forward as encoder and an LSTM network as a decoder and perform well in the patent dataset.  
\section{Conclusion}
In this paper, we introduce a model named HMT for assign the proposal to a hierarchical discipline system. We first propose a two-level hierarchical encoder to capture the representation of the proposal fully. Then, we use a Transformer-based decoder incorporating the previous label prediction and the proposal's representation to generate the label. Due to the characteristic of the decoder, our model can start from the root label or any given previous label, which corresponds to utilize the expert knowledge. We conduct extensive experiments to evaluate our model performance. The results show that our model achieves the best performance in the HMPC problem and is effective for the utilization of expert knowledge. Other experiments evaluate the ability of our model to give a proper granularity and to capture the label dependency.

\bibliographystyle{IEEEtran}
\bibliography{main}
\end{document}